\documentclass[10pt]{article} % For LaTeX2e
\usepackage[accepted]{tmlr}
\usepackage{amsmath,amsfonts,bm}

\def\eqref#1{equation~\ref{#1}}
\def\1{\bm{1}}

\DeclareMathAlphabet{\mathsfit}{\encodingdefault}{\sfdefault}{m}{sl}
\SetMathAlphabet{\mathsfit}{bold}{\encodingdefault}{\sfdefault}{bx}{n}

\usepackage{hyperref}
\usepackage{url}
\usepackage{booktabs}
\usepackage{amsmath}
\usepackage{graphicx}
\usepackage{wrapfig}
\usepackage{enumitem}

\title{Language Models for Controllable DNA Sequence Design}

\author{\name Xingyu Su \email xingyu.su@tamu.edu \\
    \addr Texas A\&M University
    \AND
    \name Xiner Li \email lxe@tamu.edu \\
    \addr Texas A\&M University
    \AND
    \name Yuchao Lin \email kruskallin@tamu.edu\\
    \addr Texas A\&M University 
    \AND
    \name Ziqian Xie \email ziqian.xie@uth.tmc.edu\\
    \addr University of Texas Health Science Center at Houston
    \AND
    \name Degui Zhi \email degui.zhi@uth.tmc.edu\\
    \addr University of Texas Health Science Center at Houston
    \AND
    \name Shuiwang Ji\thanks{Corresponding author} \email sji@tamu.edu\\
    \addr Texas A\&M University
}

\def\month{12}  % Insert correct month for camera-ready version
\def\year{2025} % Insert correct year for camera-ready version
\def\openreview{\url{https://openreview.net/forum?id=itwnoEu60S}} % Insert correct link to OpenReview for camera-ready version

\begin{document}

\maketitle

\begin{abstract}
We consider controllable DNA sequence design, where sequences are generated by conditioning on specific biological properties. 
While language models (LMs) such as GPT and BERT have achieved remarkable success in natural language generation, their application to DNA sequence generation remains largely underexplored. In this work, we introduce {ATGC-Gen}, an Automated Transformer Generator for Controllable Generation, which leverages cross-modal encoding to integrate diverse biological signals.
% such as cell types, transcription factor identity, and base-resolution transcriptional activity. 
ATGC-Gen is instantiated with both decoder-only and encoder-only transformer architectures, allowing flexible training and generation under either autoregressive or masked recovery objectives. We evaluate ATGC-Gen on representative tasks including promoter and enhancer sequence design, and further introduce a new dataset based on ChIP-Seq experiments for modeling protein binding specificity. 
Our experiments demonstrate that ATGC-Gen can generate fluent, diverse, and biologically relevant sequences aligned with the desired properties. Compared to prior methods, our model achieves notable improvements in controllability and functional relevance, highlighting the potential of language models in advancing programmable genomic design. The source code is released at (\url{https://github.com/divelab/AIRS/blob/main/OpenBio/ATGC_Gen}).
\end{abstract}

\section{Introduction}
DNA sequence design is a transformative scientific endeavor that revolutionizes our understanding of biology and catalyzes major advances across healthcare, agriculture, and environmental conservation, etc~\citep{zrimec2022controlling,killoran2017generating}.
%While current research primarily focuses on understanding DNA sequences — a long-standing topic among biologists and computer scientists — limited attention is given to the task of DNA sequence generation.
In these tasks, it is commonly expected that the generated DNA sequence achieves some biological outcomes~\citep{uehara2025reward,li2025dynamic}. Thus, controllable generation is of practical significance~\citep{evo} in which the sequence generation is guided by various properties, \emph{e.g.,} binding to specific proteins or exhibiting particular transcription activation likelihoods. 

% Recent studies on control
Recent advances in generative models have triggered widespread interests in DNA sequence design using
diffusion~\citep{avdeyev2023dirichlet,discdiff,sarkar2024designing} and flow matching~\citep{stark2024dirichlet} generative methods. These methods have shown promise, particularly in modeling global structure and optimizing over continuous latent spaces. However, they are not inherently tailored to discrete, symbolic sequence generation---properties that are central to DNA sequences.  While those methods are in general effective in generation task, they are not naturally designed for generating discrete and variable-length sequences. In contrast, Language models (LMs) offer an alternative perspective: they are naturally suited for discrete, variable-length generation and have achieved remarkable success in analogous domains like natural language. In this work, we explore the potential of LMs for controllable DNA sequence design, proposing them as a complementary approach to existing generative methods.

% Early works propose to generate DNA sequences without any controls. For example, Evo~\citep{evo} tries to generate DNA sequence at the whole genome level directly by their pretrained model. However, this kind of uncontrolled generation task is of little biological significance. Other works try to generate DNA sequences based on some requirements. \citet{avdeyev2023dirichlet} proposes a DNA promoter generation dataset, generating DNA sequence based on the transcription activation likelihood. \citet{taskiran2024cell} suggests enhancer design dataset, generating DNA enhancer sequence based on cell types. Following this, we propose to establish a DNA generation dataset based on ChIP-Seq experiment, namely, generating DNA sequences binding to specific proteins (e.g., different transcription factors) given different cell types.

In this paper, we aim to explore the use of transformer-based language models for DNA sequence design conditioned on specific biological properties~\citep{vaswani2017attention,stark2024dirichlet,sarkar2024designing,su2025learning}. To this end, we develop ATGC-Gen, an \textbf{A}utomated \textbf{T}ransformer \textbf{G}enerator for \textbf{C}ontrollable \textbf{Gen}eration. 
We instantiate ATGC-Gen with both decoder-only (e.g., GPT) and encoder-only (e.g., BERT) transformer architectures to examine their respective capabilities for controllable generation. 
It is designed to encode and integrate heterogeneous biological information, such as cell types, protein sequences, and transcription activation signals, into the sequence generation process. This unified framework enables ATGC-Gen to capture complex relationships between DNA sequences and their biological contexts, facilitating the generation of sequences with specific desirable properties.

% To evaluate the performance of ATGC-Gen, we first use the existing generation datasets in promoter~\citep{avdeyev2023dirichlet,stark2024dirichlet,sarkar2024designing} and enhancer~\citep{taskiran2024cell}. To fully explore the power of ATGC-Gen in considering complex and realistic biological properties, we develop a new controllable DNA generation dataset based on ChIP-Seq experiments. This dataset involves designing DNA sequences that bind to specific proteins, such as various transcription factors. 
% % The dataset consists of 162,642 samples in which each sample captures the binding relations between a DNA sequence and a specific protein within a specific cell type. 
% % We propose to evaluate the generation quality using three metrics that measure functionality, fluency, and diversity. Functionality measures whether the generated sequences achieve the desired binding outcomes to the proteins. Fluency assesses the naturalness and coherence of the sequences, and diversity measures the variability of the generated sequences. 
% % We conduct extensive experiments on the new ChIP-Seq dataset along with the existing promoter and enhancer datasets. 
% Our experimental results demonstrate the exceptional generative capability of ATGC-Gen, highlighting its potential for advancing DNA sequence generation under various biological property properties. 
% % We plan to release our datasets and source code upon publication.

To evaluate the effectiveness of our proposed approach, we apply ATGC-Gen, a transformer-based language model that incorporates cross-modal biological information for controllable DNA sequence generation. 
By conditioning on diverse properties such as cell types, transcription factors, and regulatory signals, ATGC-Gen enables flexible and biologically grounded sequence design. We first assess its performance on established promoter~\citep{avdeyev2023dirichlet,stark2024dirichlet,sarkar2024designing} and enhancer~\citep{taskiran2024cell} generation tasks. 
To further explore its capability in handling complex biological contexts, we introduce a new dataset based on ChIP-Seq experiments, which involves generating sequences that bind to specific proteins in specific cell types. 
Experimental results across different tasks, evaluated in terms of functionality, fluency, and diversity, demonstrate that ATGC-Gen achieves strong and consistent performance, highlighting the promise of language models for advancing controllable and property-aware DNA sequence design.

In summary, our main contributions are as follows:
\begin{itemize}[left=0in]
    \item We propose \textbf{ATGC-Gen}, a language model framework for controllable DNA sequence generation. ATGC-Gen supports flexible conditioning on diverse biological modalities, enabling biologically meaningful sequence design under complex control constraints.
    \item We introduce a new dataset for controllable DNA generation based on {ChIP-Seq experiments}, which captures protein-DNA binding patterns. The benchmark includes well-structured evaluation metrics to assess functionality, fluency, and diversity.
    \item We conduct extensive experiments on promoter, enhancer, and ChIP-Seq-based tasks. Results demonstrate that {ATGC-Gen outperforms strong baselines} in generating accurate, coherent, and diverse sequences under various biological conditions.
\end{itemize}
% 1. We propose ATGC-Gen, the first language model based framework for controllable DNA sequence generation. ATGC-Gen establishes the relationship between DNA and other domains, enabling the generation of DNA sequences according to the properties from various modalities.

% 2. We introduce a dataset derived from ChIP-Seq experiments, mapping protein sequences to DNA sequences. This dataset is of significant biological importance, providing a realistic and meaningful context of DNA generation tasks.

% 3. We conduct extensive experiments across different DNA generation tasks. The results show the superior capability of ATGC-Gen in generating DNA sequences under diverse properties, highlighting its effectiveness and potential applications.

\section{Background and Related Work}

\noindent\textbf{DNA Generation.}
We study the problem of DNA sequence generation.
% is the synthesis of DNA sequences given some certain requirements. 
In the field of synthetic biology and genetic engineering, unconditional or random DNA generation has limited applications, since DNA sequences are designed or edited to achieve specific outcomes, \emph{e.g.,} producing a protein or altering a metabolic pathway. An unconditionally generated DNA sequence lacks the necessary context to be meaningful or useful in biological systems.
Hence, in this work, we focus on studying the task of DNA generation given specific properties, such as a particular organism, cell type, or functional requirement.
% Building on these methodologies, we can formulate our research problem below. 
% \textcolor{blue}{
Formally, we define a DNA dataset as
$\mathcal{U} = \{(U_1, S_1), (U_2, S_2), \cdots, (U_N, S_N)\}$, where $U_i$ is a DNA sequence and $S_i$ is its corresponding biological property.
We aim to learn a conditional generative model $p_\theta(\cdot|S_i)$ on $\mathcal{U}$ so that the model $F_{\theta}$, parameterized with $\theta$, can generate DNA sequences fulfilling $S_i$.
% }
% Formally, given a set of biological properties $cons$, our goal is to generate DNA sequences 
% fulfilling the properties, i.e., 
% \begin{equation}
%     \{u_1,\cdots,u_n\} = \mathcal{G}_{\theta}(cons),
% \end{equation}
% where $\mathcal{G}_{\theta}$ is the generation model, and the properties $cons$ consists of various factors, including the type of DNA class, its functional attributes, and the composition of properties.
Previous studies have made several initial attempts to apply deep learning methods for controllable DNA generation. The Dirichlet Diffusion Score Model (DDSM)~\citep{avdeyev2023dirichlet} uses Dirichlet distribution to discretize the diffusion process, effectively generating DNA sequences. In \cite{stark2024dirichlet} the Dirichlet distribution is used in the flow matching process, enhancing the quality of generated DNA sequences. DiscDiff~\citep{discdiff} employs Latent Discrete Diffusion model on the DNA generation task and proposes a post-training refinement algorithm to improve the generation quality. D3~\citep{sarkar2024designing} uses score entropy for discrete diffusion on the conditional DNA sequence generation.

% While those methods achieve good results at generating DNA sequences, language models can be a more natural choice at generating discrete sequence data. 

% \subsection{Problem Formulation}
% DNA is a fundamental component in the field of biology. In this paper, given specific biological properties $cons$, we aim to generate DNA sequences $\{u_1,\cdots,u_n\}$ satisfying the requirements, i.e., 
% \begin{equation}
%     \{u_1,\cdots,u_n\} = \mathcal{G}_{\theta}(cons),
% \end{equation}
% where the properties $cons$ can vary from different modalities, such as class type and DNA functionality, and composition of properties. 

\noindent\textbf{DNA Language Model.}
Recent research has demonstrated the power of language models for modeling biomolecular discrete sequences~\citep{wang2024lc,sgarbossa2025protmamba,xu2024smiles}, which is ideally suited for modeling DNA sequences. A few recent studies have employed language models for encoding DNA sequences in prediction or representation learning tasks. Early works adopt Transformers to encode DNA sequence data, and notable examples include DNABERT, DNABERT-2 and Nucleotide Transformer~\citep{dnabert, dnabert2, NT}. More recently, state space models such as HyenaDNA~\citep{hyenaDNA} and Mamba~\citep{gu2023mamba} have been applied to long DNA sequences, with Caduceus~\citep{schiff2024caduceus} further improving Mamba with bi-directional encoding and reverse complement awareness. Evo~\citep{evo} extends this line of work by pretraining on DNA, RNA, and protein modalities. In parallel, Bio-xLSTM~\citep{schmidinger2024bio} explores xLSTM architectures for biological sequences, which, similar to state space models, achieve a linear runtime dependency on sequence length and provide efficient sequence modeling. These approaches primarily propose efficient modeling architectures for biological sequences and have achieved strong results in prediction benchmarks. 
While powerful for encoding and prediction, these works do not directly address explicit controllable sequence design. In practice, although uni-modal models can achieve limited controllability by conditioning on sequences with desirable properties, generating condition-specific DNA sequences typically requires re-training or fine-tuning a separate model for each condition. By contrast, our framework integrates heterogeneous biological properties as conditioning inputs, providing a unified and scalable way to support diverse design criteria.

% }

% \textcolor{blue}{
\noindent\textbf{Cross-Modality Representation Learning in Biology.}
Many works combine heterogeneous modalities in latent spaces in biomolecular modeling. 
For example, some works~\citep{schimunek2023context,fifty2023context,zhou2025multimodal} fuse molecular structure with other modality information, achieving strong performance on molecular property prediction benchmarks. 
Other efforts integrate different modalities of protein information to improve representation learning~\citep{gligorijevic2021structure,brandes2022proteinbert,hayes2025simulating}. 
However, these approaches focus primarily on prediction tasks rather than sequence generation, and the modalities considered are typically molecular or protein structure and activity labels. 
In contrast, our work extends cross-modal conditioning to the setting of \emph{controllable DNA sequence generation}, where diverse biological attributes are incorporated as conditioning signals to guide the design of functional DNA sequences.
% }

\section{The Proposed ATGC-Gen}

\begin{figure}
    \centering
    \includegraphics[width=\textwidth]{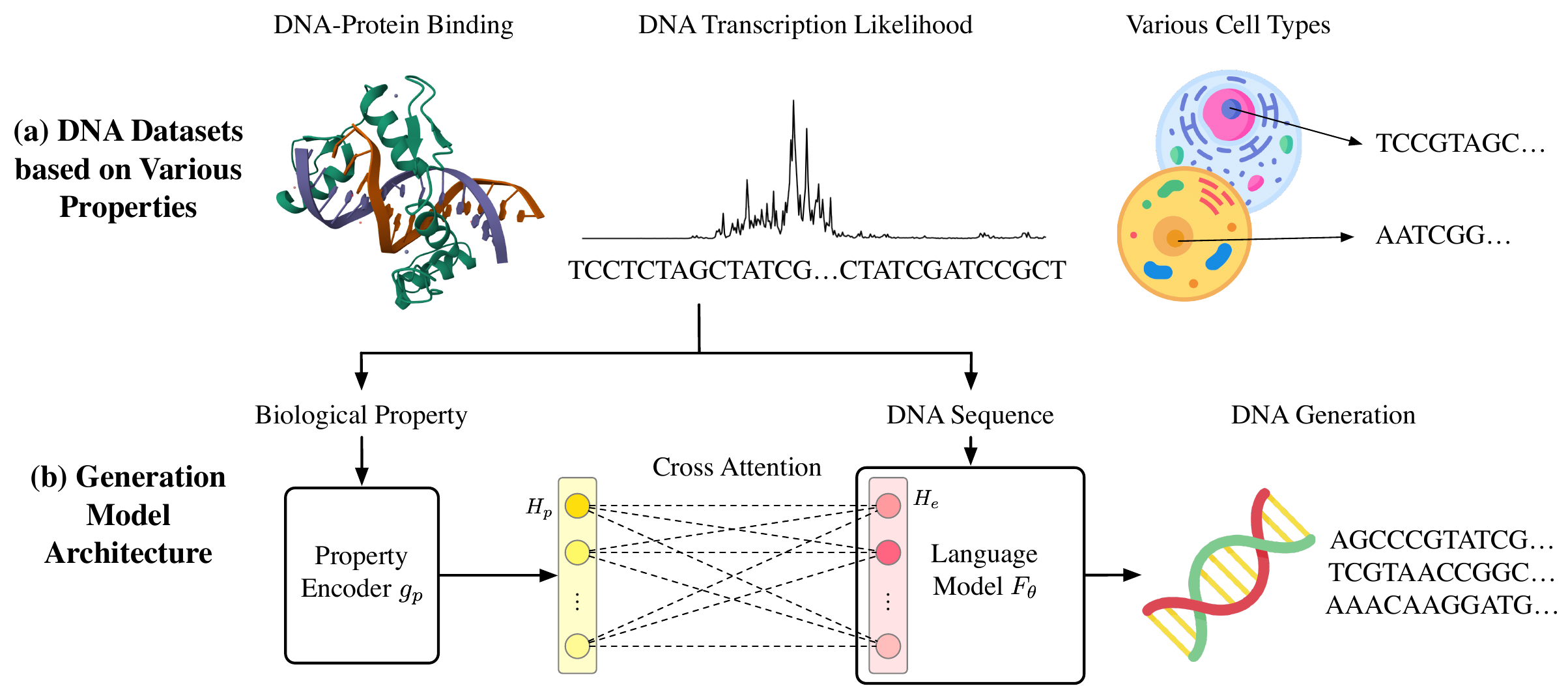}
    \caption{Overview of the proposed ATGC-Gen framework. (a) Different biological properties used for generating DNA sequences. (b) The architecture of the generation model.}
    \label{framework}
    % \vspace{-0.3cm}
\end{figure}

In this section, we introduce the overall framework of \textbf{ATGC-Gen} for controllable DNA sequence generation, as illustrated in Figure~\ref{framework}. The framework encodes various biological property modalities and employs a language model to generate DNA sequences that are aligned with specified properties. We begin by describing how the property encoder is used to transform biological property inputs into dense representations. Next, we explain how these property representations are integrated with the DNA sequence embeddings to condition the language model. Finally, we present the training objective used to optimize the model.

\subsection{Representation Encoding}
\label{sec:represent}
In this section, we illustrate encoding method for target properties from different modalities.

\paragraph{Sequence-level Integration.}
Sequence-level integration provides a general mechanism for incorporating biological properties in a global form. This approach encodes the property as a set of global representations that summarize the desired biological context---such as cell types or pooled statistics.

Given a property matrix $\mathbf{P} \in \mathbb{R}^{l_p \times d_p}$, where $l_p$ is the number of global property tokens and $d_p$ is the input feature dimension, we transform it into the model's hidden space using a learnable linear projection $\mathbf{H}_p = \mathbf{P} W_p^\top + \mathbf{b}_p$,
% where $W_p \in \mathbb{R}^{d_p \times d_h}$ and $\mathbf{b}_p \in \mathbb{R}^{d_h}$, resulting in $\mathbf{H}_p \in \mathbb{R}^{l_p \times d_h}$.
where $W_p$ and $b_p$ are learnable parameters. 

Let the DNA embedding sequence be $\mathbf{H}_e = \{\mathbf{h}_{e,1}, \cdots, \mathbf{h}_{e,n}\} \in \mathbb{R}^{n \times d_h}$, where each $\mathbf{h}_{e,i}$ is obtained by mapping the $i$-th DNA token through a learnable embedding layer that projects discrete nucleotides into a $d_h$-dimensional continuous space. We prepend the property embedding to the DNA sequence to form the final transformer input:
\begin{equation}
    \mathbf{H}_{\text{input}} = \{\mathbf{H}_p; \mathbf{H}_e\} \in \mathbb{R}^{(l_p + n) \times d_h}.
\end{equation}

This design allows the transformer to treat the property tokens as global control inputs, enabling them to influence generation via self-attention across the full input sequence.

\paragraph{Feature-level Integration.}
Feature-level integration can be applied when the property sequence is aligned position-wise with the DNA sequence, i.e., both have the same length. This setting is applicable to base pair resolution properties, such as transcriptional signals defined at each nucleotide position.

Let the DNA sequence be represented as a token sequence $U = \{u_1, \cdots, u_n\}$, where $u_i \in \{A, C, G, T\}$ representing the specific nucleotide. We convert it to one-hot embeddings $\mathbf{X}_e = \{\mathbf{x}_{e,1}, \cdots, \mathbf{x}_{e,n}\}$, where $\mathbf{x}_{e,i} \in \mathbb{R}^4$ is the one-hot encoding of $u_i$.
Let the aligned property sequence be $\mathbf{S} = \{\mathbf{s}_1, \cdots, \mathbf{s}_n\}$ with each $\mathbf{s}_i \in \mathbb{R}^{d_s}$.

For each position $i$, we concatenate the DNA one-hot and property vectors:
% \textcolor{blue}{
\[
\tilde{\mathbf{x}}_i = \text{concat}(\mathbf{x}_{e,i}, \mathbf{s}_i) \in \mathbb{R}^{(4 + d_s) \times 1}.
\]
% }
We then apply a shared linear transformation to map the concatenated vector to the model's hidden dimension:
\begin{equation}
\label{Eq:hei}
\tilde{\mathbf{h}}_{e,i} = W_f \tilde{\mathbf{x}}_i + \mathbf{b}_f,
\end{equation}
where $W_f \in \mathbb{R}^{d_h \times (4 + d_s)}$ and $\mathbf{b}_f \in \mathbb{R}^{d_h}$. 
The resulting transformer input is
\begin{equation}
    \mathbf{H}_{\text{input}} = \{\tilde{\mathbf{h}}_{e,1}, \cdots, \tilde{\mathbf{h}}_{e,n}\} \in \mathbb{R}^{n \times d_h}.
\end{equation}

Feature-level integration enables the model to condition on fine-grained biological signals at each position in the DNA sequence.

\subsection{Training Objectives}

We consider two alternative training paradigms for controllable DNA sequence generation: an \textit{autoregressive} training objective based on decoder-only transformers (e.g., GPT), and a \textit{masked language model} objective based on encoder-only transformers (e.g., BERT). Both approaches condition on external biological properties as described in the previous section.

\paragraph{Autoregressive Training.}

% \textcolor{blue}{
In this setting, we train a decoder-only language model $F_\theta$ to generate DNA sequences autoregressively.
To unify notation, let $X_i$ denote the model input at step $i$: 
(i) under \emph{sequence-level integration}, $X_i=\{\mathbf{H}_{p}; u_1,\ldots,u_i\}$, where the prepended property tokens act as a global prefix; 
(ii) under \emph{feature-level integration}, $X_i=\{\tilde{\mathbf{h}}_{e,1},\ldots,\tilde{\mathbf{h}}_{e,i}\}$, where $\tilde{\mathbf{h}}_{e,j}$ is defined in~\eqref{Eq:hei}. 
% }

Following standard practice~\citep{dnabert2,schiff2024caduceus,radford2018improving}, we use a single-nucleotide tokenizer, where each base (A, C, G, T) is treated as a discrete token. The model is trained using the next-token prediction objective with cross-entropy loss

% \textcolor{blue}{
\begin{equation}
\min_\theta \,\, \mathbb{E}_{U \sim \mathcal{U}} \left[ \sum_{i=1}^{|U|-1} \ell \big( F_\theta(X_i),\, u_{i+1} \big) \right],
\end{equation}
% }
where $U = \{u_1, \cdots, u_{|U|}\}$ is a DNA sequence from the dataset $\mathcal{U}$ and $\ell$ is the cross-entropy loss over the predicted nucleotide distribution. 

This autoregressive formulation naturally supports variable-length sequence generation, making it suitable
for DNA design tasks with flexible output lengths.

% In this setting, we train a decoder-only language model $F_\theta$ to generate DNA sequences in an autoregressive way, conditioned on the property representations $\mathbf{H}_{{p}}$. The input sequence consists of the property tokens followed by the DNA tokens, and the model is trained to predict the next nucleotide at each position. This training paradigm naturally complements {sequence-level integration}, where the prepended property tokens act as a global prefix that controls all subsequent predictions.

% Following standard practice~\citep{dnabert2,schiff2024caduceus,radford2018improving}, we use a single-nucleotide tokenizer, where each base (A, C, G, T) is treated as a discrete token. The model is trained using the next-token prediction objective with cross-entropy loss:
% \begin{equation}
% \min_\theta \,\, \mathbb{E}_{U \sim \mathcal{U}} \left[ \sum_{i=1}^{|U|-1} \ell \left( F_\theta\left(\mathbf{H}_{p}, u_1, \cdots, u_i \right), u_{i+1} \right) \right],
% \end{equation}
% where $U = \{u_1, \cdots, u_{|U|}\}$ is a DNA sequence from the dataset $\mathcal{U}$, and $\ell$ is the cross-entropy loss over the predicted nucleotide distribution.

% This autoregressive formulation naturally supports variable-length sequence generation, making it suitable for DNA design tasks with flexible output lengths.

\paragraph{Masked Language Modeling.}
In parallel, we consider an encoder-based training approach using masked language modeling (MLM), similar to BERT~\citep{devlin2019bert}. 
This formulation requires sequences of fixed length and is trained to recover randomly masked tokens in the input. 
% \textcolor{blue}{
Specifically, given an input embedding sequence $\mathbf{H}_{\text{input}}$ (constructed from both property and DNA tokens as described in Section~\ref{sec:represent}), 
% }
we randomly select a subset of positions $\mathcal{M} \subset \{1, \cdots, n\}$ to mask, and train the model $F_\theta$ to reconstruct the original nucleotides at those positions:
\begin{equation}
\min_\theta \,\, \mathbb{E}_{U \sim \mathcal{U}} \left[ \sum_{i \in \mathcal{M}} \ell \left( F_\theta\left(\mathbf{H}_{\text{input}}\right)_i, u_i \right) \right],
\end{equation}
where $\mathbf{H}_{\text{input}}$ is the full sequence embedding (property and DNA sequence with some tokens masked), and $F_\theta(\cdot)_i$ denotes the model prediction at position $i$.

Although the masked language model objective requires fixed-length sequences and is not naturally suited for generation, it enables efficient parallel training and leverages bidirectional context, making it expressive and effective in some certain cases.

\subsection{Controllable Generation}

Given a trained model and specified biological properties, we perform controllable DNA sequence generation by conditioning on the encoded property representations described earlier. We support two generation strategies, corresponding to the two training objectives: autoregressive decoding with a decoder-only (GPT-style) model, and masked language modeling with an encoder-only (BERT-style) model.

\paragraph{Autoregressive Generation.}
In the GPT-style setup, generation proceeds from left to right. At each time step $i$, the model predicts the next nucleotide $u_i$ conditioned on the property representation $\mathbf{H}_p$ and previously generated tokens $u_1, \cdots, u_{i-1}$:
\[
p_\theta(u_i \mid \mathbf{H}_p, u_1, \cdots, u_{i-1}).
\]
We initialize generation with a special start token and stop when a designated end-of-sequence token is generated or a maximum sequence length is reached.

To promote diversity and controllability, we sample tokens using temperature sampling~\citep{ackley1985learning, ficler2017controlling}.

\paragraph{Masked Recovery Generation.}
For the BERT-style model, generation is performed by iterative or parallel masked token prediction. We initialize the entire sequence with masked tokens: $\{[\texttt{MASK}], \cdots, [\texttt{MASK}]\}$, and jointly condition on the property representation $\mathbf{H}_p$. The model then predicts the nucleotide for each masked position randomly.

There are two generation modes in this setting:
\begin{itemize}[left=0in]
    \item \textbf{One-shot unmasking:} all positions are predicted in parallel from a fully masked input in a single forward pass.
    \item \textbf{Iterative unmasking:} a subset of masked positions is predicted at each step, and the predictions are fed back in; the remaining positions stay masked and are updated progressively.
\end{itemize}

This formulation enables parallel generation and full bidirectional conditioning, but requires fixing the sequence length in advance.

\textit{Remark.} This formulation shares structural similarities with recent masked denoising approaches, such as masked discrete diffusion~\citep{sahoo2024simple,nie2025large}, where a corrupted input is progressively reconstructed through iterative masked token prediction.

\section{Controllable DNA Generation with ChIP-Seq}
\label{chipseq_data}

\begin{table}
    \caption{Examples of the ChIP-Seq generation dataset.}
    \label{chipseq_example}
    \centering
    \begin{tabular}{cccccc}
    \toprule
    Chrom & ChromStart & ChromEnd & Transcription Factor (TF) & Cell Type & Score \\
    \midrule
    % chrY & 2657879 & 2658063 & CTCF & 652 & GM19239 \\
    chr1 & 26677454 & 26677790 & TCF7 & K562 & 405 \\
    % chr16 & 88717317 & 88717773 & CCNT2 & 132 & K562 \\
    chr4 & 64158376 & 64159457 & CTCF & medulloblastoma & 1000 \\
    % chr12 & 48730360 & 48730650 & ZNF143 & 194 & H1-hESC \\
    chr2 & 190070290 & 190070667 & RAD21 & GM12878 & 1000 \\
    chr11 & 86800296 & 86800780 & MXI1 & SK-N-SH & 277 \\
    % chr8 & 146277609 & 146278073 & MAZ & 701 \\
    % chr7 & 577857 & 578321 & MAZ & 262 \\
    \bottomrule
    \end{tabular}
    % \vspace{-0.3cm}
\end{table}

% \textcolor{blue}{
While existing DNA datasets are sufficient to benchmark the DNA generation tasks, they do not capture the variable-length nature of many real biological sequences. 
We therefore curate a variable-length dataset to highlight the adaptability of our framework to more realistic and flexible generation scenarios.
% }
Therefore, we develop a dataset derived from ChIP-Seq experiments for evaluating controllable DNA generation problems. ChIP-Seq experiments aim to identify the binding sites of DNA-associated proteins on the genome, specifically the DNA sequences where proteins physically attach or bind. In our setting, we focus on predicting potential binding DNA sequences of variable lengths based on given protein sequences across different cell types. This prediction can advocate our understanding of the complex networks of gene regulation and aid in designing DNA sequences with specific functionalities for genetic engineering.

Our proposed dataset is obtained from a comprehensive collection of ChIP-Seq experiments generated by the ENCODE project~\citep{encode2012integrated}, which identifies specific DNA sequences to which proteins bind. We filter the raw data by considering the highest binding scores, length properties and other criteria, and detailed steps are provided in the Appendix~\ref{app_data}. 

\noindent\textbf{Dataset Description.}
Table~\ref{chipseq_example} shows several example rows of the ChIP-Seq generation dataset. The raw dataset contains approximately 10 million rows, with each row describing a binding relationship between a DNA sequence and a protein for specific cell type(s). This dataset contains 340 transcription factors (proteins) and 129 cell types. In addition to the (DNA, protein, cell type) triplets, each row presents the binding scores to indicate the binding strength. 

The DNA sequences are from GRCh38 (HG38) reference human genome assembly, used for mapping and aligning genetic data. Each base pair in the DNA sequence is one of the four nucleotides: adenine (A), cytosine (C), guanine (G), thymine (T). Given the protein name, we can query protein databases~\citep{PDB} to find the corresponding protein sequences and structures.

% \noindent\textbf{Controlled Generation Task}
% 

\begin{figure}
    \centering
    \includegraphics[width=\textwidth]{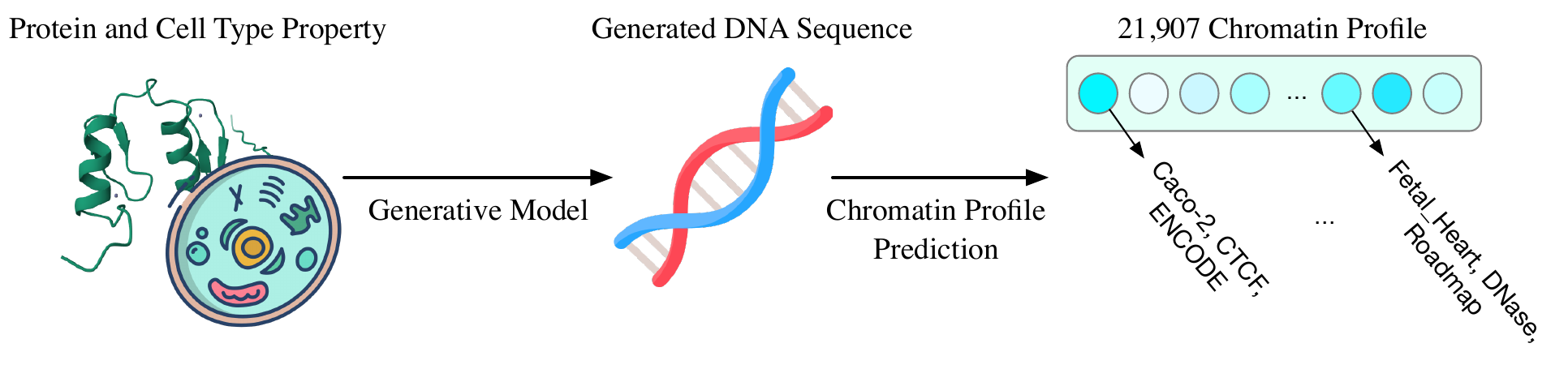}
    \caption{Illustration of the ChIP-Seq generation task. The objective is to generate DNA sequences based on specific proteins and cell types. The functionality of the generated DNA sequences is evaluated using chromatin profile prediction.}
    \label{dataset}
    % \vspace{-0.3cm}
\end{figure}
\noindent\textbf{Task and Evaluation Metrics.}
The controlled generation task involves generating DNA sequences based on a given cell type and binding protein. Figure~\ref{dataset} illustrates the generation task. 
% Here, the properties are both the cell types and protein sequences. 
It is important to note that for the DNA-protein binding, a single protein may bind with different DNA sequences, and the same DNA sequence can bind to different proteins.

We propose three quantitative metrics to evaluate the generated sequences.
\begin{itemize}[left=0in]
    \item \textbf{Functionality}: This metric measures how well the generated DNA sequence binds with the given protein under a specific cell type. We use the SEI framework~\citep{SEI}, which employs a deep learning model to predict 21,907 chromatin profiles based on DNA sequences. As shown in Figure~\ref{dataset}, each chromatin profile is defined by cell type, protein name and experiment. The prediction value indicates the probability of binding between the DNA and the protein. We focus exclusively on chromatin profiles derived from ENCODE experiments, as our ChIP-Seq dataset is sourced from the ENCODE project. 
    
    \item \textbf{Fluency}: Similar to the fluency used in the text generation task~\citep{diffusionLM,instructCTG}, we adopt it to assesses how smooth and natural the generated DNA sequence is. Fluency is defined as the perplexity of the generated DNA in a pretrained language model, with lower value indicating better fluency. We use the HyenaDNA pretrained model as the base model $F_b$~\citep{hyenaDNA}. Given the cross-entropy loss $\ell$, the fluency of a single DNA sequence $U=\{u_1,\cdots,u_n\}$ is the exponential of the loss
    \begin{equation}
        \text{Flu} = \exp( \sum_{i} \ell \left( F_b (u_1, \cdots, u_i), u_{i+1}\right) ).
    \end{equation}
    The total fluency is then calculated as the mean fluency across all DNA sequences.

    \item \textbf{Diversity}: In a specific cell type, a given protein can bind to many different DNA sequences. This metric measures the diversity of the generated sequences. Following \cite{discdiff}, the diversity score is defined as
    \begin{equation}
        \text{Diversity}= \sum_c (W_c \times \prod_{n=10}^{12} \frac{|\text{count}(n,U)|}{\text{count}(n,U)}),
    \end{equation}
    where $\text{count}(n,U)$ is the total number of n-grams in the generated DNA sequence $U$, and $|\text{count}(n,U)|$ is the number of unique n-grams. Given the variability in the number of samples in each category $c$, the weight of diversity in each category is denoted by $W_c=N_c/N$, where $N_c$ is the number of samples in category $c$ and $N$ is the total number of samples.
\end{itemize}

\section{Experiments}
% To demonstrate the effectiveness of our proposed ATGC-Gen, we conduct experiments on three datasets below.

\subsection{Promoter Generation}
\label{Exp:promoter}
\noindent\textbf{Dataset and task descriptions.}
This dataset contains 100,000 promoter sequences from GRCh38 (HG38) human reference genome. Each sequence is annotated with the transcription initiation profile, indicating the likelihood of transcription initiation activity at each base pair \citep{fantom2014promoter}. We use the same DNA sequences as in previous works \citep{avdeyev2023dirichlet, stark2024dirichlet}, specifically splitting 1,024 base pair long DNA sequences around each annotated transcription start site position \citep{hon2017atlas}. The task is to interpret the transcription initiation profile into the corresponding DNA sequences of the same length.

\textbf{Metrics.}
Following prior works~\citep{avdeyev2023dirichlet, stark2024dirichlet}, we evaluate the generation performance using the mean squared error (MSE) between the predicted regulatory activity of the generated and original DNA sequences. The regulatory activity is computed using the SEI framework~\citep{SEI}, a published deep learning model that predicts 21,907 regulatory features for a given DNA sequence. For our primary evaluation, we extract promoter-related activity by focusing on the H3K4me3 chromatin mark.
In addition, we report the Kolmogorov–Smirnov (KS) statistic between the distributions of regulatory activity values from the generated and original sequences, as proposed in~\citep{sarkar2024designing}, to assess distributional alignment.
We further evaluate the fluency of the generated sequences by HyenaDNA~\citep{nguyen2023hyenadna}, as described in Section~\ref{chipseq_data}, to ensure that generated DNA is biologically plausible and structurally coherent.

\begin{table}
    % \vspace{-0.4cm}
    \caption{Performance on promoter design. The best performance is indicated in bold, while the second-best performance is underlined. This convention is followed in all subsequent tables.}
    \label{promoter}
    \centering
    % \resizebox{\textwidth}{!}{
    \begin{tabular}{l|ccc}
        \toprule
        Method & MSE \textdownarrow & KS Test Statistic \textdownarrow & Fluency \textdownarrow \\
        \midrule 
        BIT Diffusion & 0.0395 & - & - \\
        D3PM-UNIFORM & 0.0375 & - & - \\
        DDSM & 0.0334 & - & - \\
        Dirichlet FM & 0.0269 & 0.399 & {3.6538} \\
        D3 & \underline{0.0219} & {0.052} & \underline{3.5255} \\
        \midrule 
        ATGC-Gen-GPT & {0.0289} & \underline{0.048} & {\textbf{3.5231}} \\
        ATGC-Gen-Bert & \textbf{0.0192} & \textbf{0.043} & {{3.5904}} \\
        \bottomrule
    \end{tabular}
    % }
\end{table}

\textbf{Results and analysis.} Table~\ref{promoter} shows the generation performances for the promoter sequences. 
The results show that \textbf{ATGC-Gen-BERT} achieves the best overall performance compared to previous baselines, demonstrating the strong modeling capacity of the masked recovery generation paradigm. In contrast, \textbf{ATGC-Gen-GPT} aggregates base-resolution activity features at the beginning of the sequence, which removes per-token alignment between the activity signals and the DNA sequence. This loss of fine-grained information leads to inferior performance in functional metrics. However, due to its autoregressive nature, ATGC-Gen-GPT better preserves the fluency and coherence of the generated DNA sequences.

\subsection{Enhancer Generation}
\noindent\textbf{Dataset and task descriptions.}
This task involves two distinct datasets: one from fly brain~\citep{flybrain} and the other from human melanoma cells~\citep{melanoma}. We follow the same data split as in the work by~\cite{stark2024dirichlet}, using 104k fly brain DNA sequences and 89k human melanoma DNA sequences, each consisting of 500 base pairs. The cell class labels are derived from the ATAC-seq experiments~\citep{buenrostro2013transposition}, with 47 classes for the human melanoma dataset and 81 classes for the fly brain dataset. Our objective is to generate DNA sequences based on the given cell class labels. 

\textbf{Metrics.} Following~\cite{stark2024dirichlet}, we evaluate the performance using the Fr\'echet Biological Distance (FBD) between the generated and original DNA sequences. To calculate the FBD score, we use a pretrained classifier model~\citep{stark2024dirichlet} to obtain the hidden representations and then compute the Wasserstein distance between Gaussians. Additionally, we assess performance using the diversity score and fluency score, similar to the metrics proposed in Section~\ref{chipseq_data}. These metrics provide a comprehensive evaluation of the generated DNA sequences.

\begin{table}
    \caption{Performance on enhancer generation.}
    \label{enhancer}
    \centering
    % \resizebox{\textwidth}{!}{
    \begin{tabular}{l|ccc|ccc}
    % \begin{tabular}{l|p{1.0cm}p{1.6cm}p{1.4cm}|p{1.0cm}p{1.6cm}p{1.4cm}}
        \toprule
         & \multicolumn{3}{c|}{Fly Brain} & \multicolumn{3}{c}{Melanoma} \\
         % \midrule 
        Method & FBD $\downarrow$ & Diversity $\uparrow$ & Fluency $\downarrow$ & FBD $\downarrow$ & Diversity $\uparrow$ & Fluency $\downarrow$ \\
        \midrule 
        % Dirichlet FM w/o Property & 15.2107 & \underline{\textbf{0.0697}} & 4.0847 & \underline{\textbf{5.3874}} & \underline{\textbf{0.0696}} & 3.8251 \\
        % ATGC-Gen w/o Property & \underline{\textbf{14.6412}} & 0.0572 & \underline{\textbf{4.0229}} & 8.3796 & 0.0579 & \underline{\textbf{3.6329}} \\
        % \hline\hline
        Dirichlet FM & 1.0404 & {\textbf{0.8314}} & 4.0512 & 1.9051 & {\textbf{0.8395}} & 3.7126 \\
        ATGC-Gen-GPT & {\textbf{0.5080}} & 0.8309 & {\textbf{4.0326}} & {\textbf{0.9228}} & 0.8131 & {\textbf{3.6852}} \\
        % ATGC-Gen-Mask & {\textbf{27.63}} & 0.8667 & {\textbf{3.9609}} & {\textbf{45.27}} & 0.8230 & {\textbf{3.7560}} \\
        \bottomrule
    \end{tabular}
    % \vspace{-0.1cm} 
    % }
\end{table}

\noindent\textbf{Results and analysis.}
Table~\ref{enhancer} presents the results for enhancer DNA sequence generation. Our proposed \textbf{ATGC-Gen-GPT} achieves a significantly lower Fr\'echet Biological Distance (FBD) than baseline models, indicating its effectiveness in incorporating global property information and modeling cross-modality signals to generate biologically realistic sequences. In terms of generation quality, ATGC-Gen-GPT produces more fluent and coherent sequences than the flow matching-based model, although with reduced diversity.
Additional results for \textbf{ATGC-Gen-BERT} are provided in Appendix~\ref{app:enhancer_bert}. Compared to the GPT variant, ATGC-Gen-BERT demonstrates weaker performance in capturing global property information, likely due to the limitations of the masked modeling objective in autoregressive-style generation.

\subsection{ChIP-Seq Generation}

\begin{table}
    \caption{Performance on ChIP-Seq DNA generation. "w/" indicates that DNA sequences are generated based on the biological properties, while "w/o" indicates that the properties are not provided during generation.}
    \label{chipseq_result}
    \centering
    \begin{tabular}{l|cc}
        \toprule
        Method & Binding Score $\uparrow$ & Diversity $\uparrow$ \\
        \midrule 
        Real Sequence & 0.2319 & 0.0513 \\
        \midrule
        Random Sequence & 0.0036 & - \\
        ATGC-Gen w/o Property & 0.0747 & 0.1213 \\
        \midrule
        ATGC-Gen w/ Property & \textbf{0.1176} & \textbf{0.1228} \\
        \bottomrule
    \end{tabular}
    % \vspace{-0.3cm} 
\end{table}

\noindent\textbf{Results and Analysis.}
Table~\ref{chipseq_result} reports the results on the ChIP-Seq generation task, described in Section~\ref{chipseq_data}. Since the task involves variable-length generation, we only evaluate \textbf{ATGC-Gen-GPT}. The model achieves a strong Binding Score when conditioned on biological properties. It also achieves higher diversity compared to other ablations.
The performance drop without property inputs highlights the importance of conditioning on biological context. Nonetheless, the property-free model still performs better than random sequences, suggesting that ATGC-Gen captures meaningful sequence patterns even without explicit control signals.

\section{Conclusion and Discussion}
In this paper, we present \textbf{ATGC-Gen}, a controllable DNA sequence generator that integrates diverse biological properties via cross-modal encoding and leverages language models for conditional generation. Our method enables accurate, fluent, and biologically relevant sequence design across multiple tasks. To support this direction, we also construct a new dataset from ChIP-Seq experiments, providing a realistic benchmark for DNA–protein binding generation.

\noindent\textbf{Practical Usage and Applications.}
Controllable sequence design is closely aligned with real-world needs in synthetic biology, such as promoter or enhancer engineering and transcription-factor binding site optimization. ATGC-Gen can assist researchers in generating candidate sequences that satisfy multiple design constraints before downstream experimental screening. 
Although this work does not include wet-lab validation, the proposed controllability interface aligns with how sequence constraints are applied in practical design settings.

\noindent\textbf{Limitations.}
Despite promising results, our work is constrained by limited computational resources, preventing the training of larger models. Moreover, frequent evaluations during training, especially under autoregressive settings, introduce notable computational overhead.

\noindent\textbf{Broader Impacts.}
Controllable DNA generation offers exciting opportunities for synthetic biology and genetic engineering. While such capabilities may accelerate the design of beneficial biological functions, they also raise concerns about misuse or the unintended synthesis of harmful sequences. Ensuring biological safety, responsible use, and ethical safeguards remains critical.

\noindent\textbf{Future Directions.}
Future work may explore more advanced generation techniques such as retrieval-augmented models, as well as the development of richer and biologically grounded benchmarks to drive progress in programmable genomics. Incorporating experimental feedback loops or active-learning strategies may further enhance the practical utility of controllable DNA generation.

%For future directions, we encourage researchers to explore more biologically meaningful benchmarks in the field of DNA generation tasks. The integration of property constraints should be considered, further improving the applicability and precision of synthetic DNA sequences. 

\section*{Acknowledgments}

This work was supported in part by National Institutes of Health under grant U01AG070112.

\clearpage
\bibliographystyle{tmlr}  
\bibliography{tmlr}

\clearpage

\appendix

\section{Data Processing for ChIP-Seq Experiment}
\label{app_data}
Based on the data from the ChIP-Seq experiment, we perform data preprocessing. No license is needed for the data files and database tables\footnote{\url{https://genome.ucsc.edu/license/}}. 
% The raw data comes from the ENCODE project\footnote{\url{https://genome.ucsc.edu/cgi-bin/hgTrackUi?db=hg19&g=wgEncodeRegTfbsClusteredV3}}. 
The raw data comes from the ENCODE project\footnote{\url{https://genome.ucsc.edu/cgi-bin/hgTrackUi?db=hg38&g=encRegTfbsClustered}}.
The preprocessing steps are as follows:

\noindent\textbf{Filtering by Binding Scores.}
Binding score measures how effectively DNA sequences bind to proteins. To focus on the most relevant interactions, we retain only the highest binding scores for each category. This step ensures that our analysis concentrates on the strongest and most meaningful DNA-protein interactions.

% \noindent\textbf{Data Explosion by Cell Types.}
% Initially, the dataset contains entries where different cell types are aggregated together if the proteins and DNA sequences are the same. For better analysis and modeling, we disaggregate, or "explode," the dataset by cell types. This step ensures that each cell type is treated as a separate entry, simplifying downstream processing and analysis.

\noindent\textbf{Alignment with SEI Framework.}
To ensure compatibility with the SEI framework used for downstream evaluation of DNA sequence functionality, we align the cell types and protein names in our dataset with those recognized by SEI. Specifically, we retain the GM12878 cell type, which has the largest number of entries. Additionally, we filter the dataset to include only proteins whose names are supported by the SEI framework, enabling accurate and consistent evaluation.

\noindent\textbf{Standardizing Input Sequence Lengths.}
The raw dataset contains pairs of DNA sequences and their associated binding transcription factor (TF) protein sequences, with lengths varying significantly, ranging from 50 to over 20,000 base pairs for DNA and up to over one thousand amino acids for proteins. Such variation poses challenges for modeling and computation. To ensure consistency and tractability, we filter the dataset to retain only DNA sequences shorter than 500 base pairs and protein sequences shorter than 1,000 amino acids. This standardization facilitates efficient training and stable convergence of the language models.

\noindent\textbf{Embedding Protein Sequences.}
To represent the transcription factor (TF) proteins, we use the ESM-2-3B protein language model~\citep{lin2023evolutionary} to extract contextual embeddings from the amino acid sequences. We retain the full-length output embeddings without applying any sequence-level pooling or aggregation, thereby preserving residue-level resolution for downstream tasks.

% \noindent\textbf{Removing Underrepresented Categories.}
% To ensure robust model training and evaluation, we filter out categories with fewer than 10 samples. Categories with very few samples would not have enough data points to be included in the validation or testing sets. We use a stratified splitting method to divide the dataset into training, validation, and testing sets, ensuring that each category is represented in all three subsets.

After completing the preprocessing steps, we partition the dataset by chromosome: chromosomes 20 and 21 are used for validation, while chromosomes 22 and X are held out for testing. The remaining chromosomes are used for training. The detailed statistics of the cleaned and processed dataset are provided in Table~\ref{app_chip_dataset}.

% Based on the data from Chip-Seq experiment, we first do data preprocessing on them. 
% Firstly, we explode the dataset by the cell types, since originally they collect all the cell types together if the proteins and DNA sequences are the same. I explode it on the cell type for convenience. Then, since we will evaluate the DNA sequence functionality by SEI framework, we need to align the cell type name string and the protein name string with that of SEI framework. We only keep the name that is shown in the SEI framework.

% Second, we observe that the length of the DNA sequences varies a lot, from 50 to about 20k. Such a huge differences are not good for modeling, so we decide to only keep DNA sequences at certain range. We keep the length of DNA sequences between 500 and 600 since DNA sequences are frequently at this range of length.

% Next, we consider the binding score. It measures how DNA sequences bind to proteins. We only keep the highest binding scores for each protein under each cell type. This makes us only consider the tight binding between DNA sequences and proteins.

% Finally, we filter out the category with samples less than 10, since if the number of samples is less than 10, this category will not have samples in the validation or test set. We use stratified splitting method to make each category have samples in all training, validation and testing dataset.

% After all the previous steps, the statistics of the dataset information is illustrated in Table~\ref{app_chip_dataset}.
\begin{table}[!htbp]
    \caption{ChIP-Seq dataset information after preprocessing.}
    \label{app_chip_dataset}
    \centering
    \begin{tabular}{ccc}
    \toprule
       \# rows & \# proteins & \# split samples  \\
       \midrule
       55830 & 62 & 51800/2181/1849	\\ 
       \bottomrule
    \end{tabular}
\end{table}

\section{Implementation Details}
\label{implement}
% \subsection{ATGC-Gen}
We report the hyperparameters and training configurations used in our experiments. 
For \textbf{ATGC-Gen-Bert}, we adopt a BERT-style encoder with the \texttt{bert-base} configuration: 12 layers, 12 attention heads, and a hidden size of 768. 
For \textbf{ATGC-Gen-GPT}, we use a GPT-style decoder with 16 layers, 16 attention heads, and the same hidden size of 768.
We use the AdamW optimizer with a learning rate of $1\times 10^{-4}$ and a linear warmup over the first 10\% of training epochs. Model selection is based on performance on the validation set. The batch size is adjusted to fit within GPU memory constraints.
All training and inference are conducted on NVIDIA A100-SXM-80GB GPUs.

% \subsection{Dirichlet FM}
% Regarding the baseline methods, most can be directly referenced from the original papers. However, for the ChIP-Seq dataset, since Dirichlet FM has not been tested on this dataset, we reproduce the results. We use the code of Dirchlet FM\footnote{\url{https://github.com/HannesStark/dirichlet-flow-matching}}, which is licensed under the MIT License. Similar to the Enhancer dataset in the Dirichlet FM paper, we consider the cell types in the ChIP-Seq dataset in the same manner. For protein embeddings, after obtaining the embeddings using the ESM-2 encoder, we concatenate the class embeddings with the protein embeddings to form the property embeddings, keeping the rest of the architecture unchanged. We test both CNN and Transformer models using different guidance values. Ultimately, we choose to set the guidance value to 3 and use classifier-free guidance to incorporate the property information.

\section{Additional Results}

\begin{table}
    \caption{Performance on enhancer generation without property information.}
    \label{enhancer_no}
    \centering
    \resizebox{\textwidth}{!}{
    \begin{tabular}{l|ccc|ccc}
        \toprule
         & \multicolumn{3}{c|}{Fly Brain} & \multicolumn{3}{c}{Melanoma} \\
         % \midrule 
        Method & FBD $\downarrow$ & Diversity $\uparrow$ & Fluency $\downarrow$ & FBD $\downarrow$ & Diversity $\uparrow$ & Fluency $\downarrow$ \\
        \midrule 
        Dirichlet FM w/o Property & 15.2107 & {\textbf{0.0697}} & 4.0847 & {\textbf{5.3874}} & {\textbf{0.0696}} & 3.8251 \\
        ATGC-Gen-GPT w/o Property & {\textbf{14.6412}} & 0.0572 & {\textbf{4.0229}} & 8.3796 & 0.0579 & {\textbf{3.6329}} \\
        % \hline\hline
        % Dirichlet FM & 1.0404 & {\textbf{0.8314}} & 4.0512 & 1.9051 & {\textbf{0.8395}} & 3.7126 \\
        % ATGC-Gen & {\textbf{0.8164}} & 0.8262 & {\textbf{4.0286}} & {\textbf{0.7910}} & 0.8069 & {\textbf{3.6773}} \\
        \bottomrule
    \end{tabular}
    % \vspace{-0.3cm} 
    }
\end{table}

% \subsection{Enhancer Design w/o Property.}
% Here we present the results for enhancer sequence generation without incorporating property information. Table~\ref{enhancer_no} shows that in scenarios where property information is not used, ATGC-Gen-GPT can have worse performances than the Dirichlet FM. However, in the controllable generation task, where property information is considered, we know ATGC-Gen-GPT consistently outperforms the Dirichlet FM method by a significant margin.

% The primary reason for this discrepancy is that ATGC-Gen excels at capturing and utilizing property constraints. This capability allows ATGC-Gen to generate sequences that are highly aligned with the desired biological properties, making it particularly effective for tasks requiring controllable generation. Conversely, in uncontrolled generation scenarios, where no property information is provided, ATGC-Gen cannot have such advantages. 

% In conclusion, while ATGC-Gen demonstrates superior performance in controllable generation tasks by using property information effectively, Dirichlet FM shows comparable results in uncontrolled generation tasks. This highlights the importance of selecting the appropriate generation model based on whether property constraints are a critical factor in the task.

\subsection{Enhancer Design without Property Information}
To better understand the role of conditioning signals, we follow prior work and evaluate ATGC-Gen-GPT in an uncontrolled setting, where no property information is provided during generation~\citep{stark2024dirichlet}. As shown in Table~\ref{enhancer_no}, ATGC-Gen-GPT results in a mixed performances compared with the Dirichlet FM baseline in this scenario.
This result contrasts with the controllable generation setting, where ATGC-Gen-GPT consistently outperforms Dirichlet FM by a large margin (Table~\ref{enhancer}). The discrepancy highlights that ATGC-Gen is particularly effective when it can leverage external biological properties to guide sequence generation. In the absence of such conditioning, its advantage diminishes.
Overall, these findings suggest that while Dirichlet FM remains competitive in unconditional generation, ATGC-Gen is better suited for property-conditioned tasks---precisely the setting that is most relevant in practical genomic design applications.

% \begin{table}
%     \caption{Experimental results with error bars in enhancer designs.}
%     \label{error_bar}
%     \centering
%     \begin{tabular}{l|cc}
%         \toprule
%         Method & FBD of Fly Brain & FBD of Melanoma \\
%         \midrule 
%         ATGC-Gen & $0.8124\pm0.0081$ & $0.7794\pm 0.1621$ \\
%         \bottomrule
%     \end{tabular}
% \end{table}

% \noindent\textbf{Statistical Significance.}
% For the task of enhancer DNA design, we run the generation process three times using different random seeds. The error bars, as reported in Table~\ref{error_bar}, demonstrate that the slight variations do not affect the overall conclusions regarding our proposed ATGC-Gen.

\subsection{Enhancer Design with ATGC-Gen-Bert}
\label{app:enhancer_bert}

\begin{table}
    \caption{Performance between ATGC-Gen-Bert and ATGC-Gen-GPT on enhancer generation.}
    \label{app:result_enh_bert}
    \centering
    % \resizebox{\textwidth}{!}{
    \begin{tabular}{l|ccc|ccc}
    % \begin{tabular}{l|p{1.0cm}p{1.6cm}p{1.4cm}|p{1.0cm}p{1.6cm}p{1.4cm}}
        \toprule
         & \multicolumn{3}{c|}{Fly Brain} & \multicolumn{3}{c}{Melanoma} \\
         % \midrule 
        Method & FBD $\downarrow$ & Diversity $\uparrow$ & Fluency $\downarrow$ & FBD $\downarrow$ & Diversity $\uparrow$ & Fluency $\downarrow$ \\
        \midrule 
        % Dirichlet FM w/o Property & 15.2107 & \underline{\textbf{0.0697}} & 4.0847 & \underline{\textbf{5.3874}} & \underline{\textbf{0.0696}} & 3.8251 \\
        % ATGC-Gen w/o Property & \underline{\textbf{14.6412}} & 0.0572 & \underline{\textbf{4.0229}} & 8.3796 & 0.0579 & \underline{\textbf{3.6329}} \\
        % \hline\hline
        % Dirichlet FM & 1.0404 & {\textbf{0.8314}} & 4.0512 & 1.9051 & {\textbf{0.8395}} & 3.7126 \\
        ATGC-Gen-GPT & {{0.5080}} & 0.8309 & {{4.0326}} & {{0.9228}} & 0.8131 & {{3.6852}} \\
        ATGC-Gen-Bert & {{27.63}} & 0.8667 & {{3.9609}} & {{45.27}} & 0.8230 & {{3.7560}} \\
        \bottomrule
    \end{tabular}
    % \vspace{-0.1cm} 
    % }
\end{table}

Table~\ref{app:result_enh_bert} presents a comparison between ATGC-Gen-BERT and ATGC-Gen-GPT on the enhancer generation tasks. Despite both variants using the same sequence-level integration to encode global property information, ATGC-Gen-BERT performs substantially worse than ATGC-Gen-GPT in terms of Fréchet Biological Distance (FBD).
This performance gap suggests that masked recovery generation, as used in ATGC-Gen-BERT, is less effective at capturing and utilizing global properties compared to the autoregressive decoding in ATGC-Gen-GPT. In contrast, the autoregressive nature of GPT allows for sequential incorporation of such information, leading to better alignment with the desired biological conditions.

\subsection{Effect of Unmasking Rate on ATGC-Gen-Bert}
\begin{figure}
    \centering
    \includegraphics[width=0.5\textwidth]{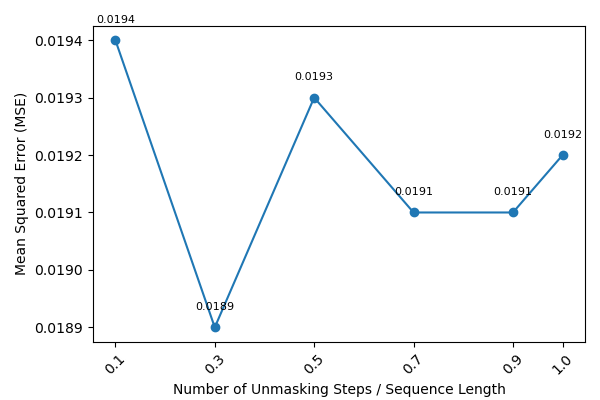}
    \caption{Effect of unmask ratio on ATGC-Gen-Bert in promoter generation.}
    \label{gen_promo_rate}
    % \vspace{-0.3cm}
\end{figure}

During the generation process of ATGC-Gen-Bert, we can control the number of tokens unmasked at each step. For instance, we may choose to unmask all masked tokens simultaneously or unmask only one randomly selected token per step. Figure~\ref{gen_promo_rate} presents the results under varying unmasking rates, where the x-axis denotes the number of unmasking steps normalized by sequence length (e.g., 1.0 corresponds to unmasking one token per step for a 1024-length sequence).
The figure shows that overall performance remains relatively stable across different unmasking schedules. Notably, using fewer steps (e.g., unmasking 10\% of tokens per step, or 103 steps in total) yields comparable performance to full-length generation (1024 steps), suggesting the potential for significant speedup. However, the extreme case of unmasking all tokens in a single step results in a notable performance drop (MSE = 0.0334), indicating the importance of stepwise refinement.
In Table~\ref{promoter}, we report results using the most conservative setting, unmasking one token per step. While not the best-performing configuration in Figure~\ref{gen_promo_rate}, we adopt this setting to avoid selecting hyperparameters based on testing set performance, which could introduce bias.

\begin{figure}
    \centering
    \includegraphics[width=1.0\textwidth]{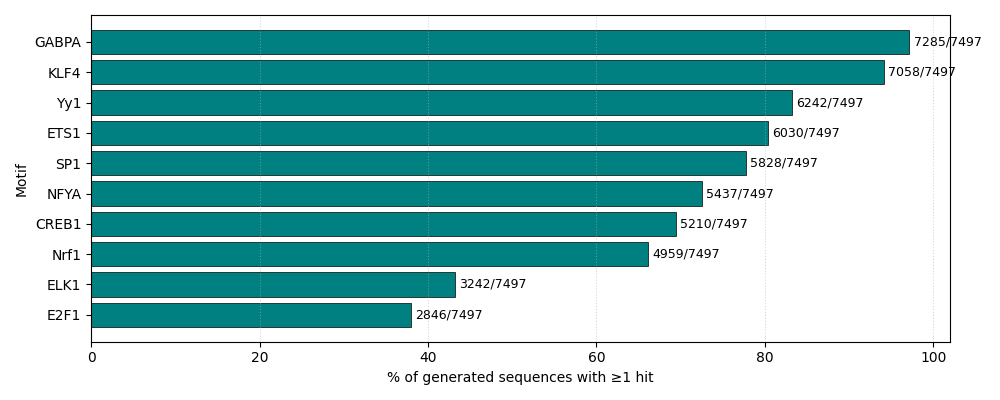}
    \caption{Fraction of generated sequences containing at least one occurrence of each selected transcription factor motif. Highly enriched motifs include GABPA, KLF4, and YY1.}
    \label{motif_present}
\end{figure}

\begin{figure}
    \centering
    \includegraphics[width=1.0\textwidth]{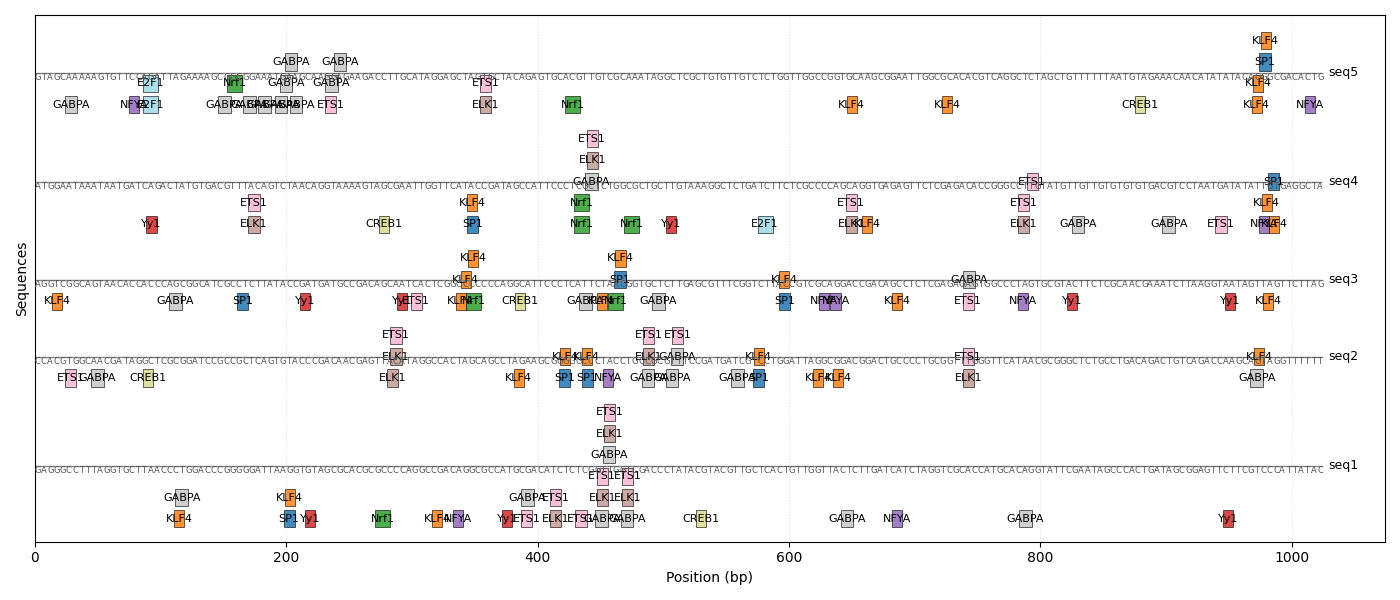}
    \caption{Motif distribution across five randomly sampled generated sequences. Each box marks a detected motif occurrence at the corresponding genomic position.}
    \label{motif_viz}
\end{figure}

% \textcolor{blue}{
\subsection{Motif Enrichment in Generated Sequences}
% }
To further validate the biological plausibility of our generated regulatory sequences, we performed motif enrichment analysis using a curated panel of transcription factor (TF) binding motifs from the JASPAR database~\citep{rauluseviciute2024jaspar} on the promoter design task described in Section~\ref{Exp:promoter}. 
Specifically, we considered ten motifs corresponding to well-established regulators of gene expression:
\begin{itemize}
    \item SP1 (MA0079.5)
    \item KLF4 (MA0039.5)
    \item NRF1 (MA0506.3)
    \item YY1 (MA0095.4)
    \item NFYA (MA0060.4)
    \item ELK1 (MA0028.3)
    \item ETS1 (MA0098.4)
    \item GABPA (MA0062.4)
    \item CREB1 (MA0018.5)
    \item E2F1 (MA0024.3)
\end{itemize}
We scanned all generated sequences for the presence of these motifs using position frequency matrices (PFMs) from JASPAR, followed by enrichment quantification.
\paragraph{Global motif enrichment.}
Figure~\ref{motif_present} summarizes the fraction of generated sequences that contain at least one occurrence of each TF motif. Highly enriched motifs include GABPA, KLF4, and YY1, with more than $80\%$ of sequences exhibiting at least one instance. This demonstrates that the model captures essential cis-regulatory elements commonly associated with enhancer-like activity.

\paragraph{Motif positional distribution.}
To illustrate the positional organization of motifs, Figure~\ref{motif_viz} shows motif annotations across five randomly selected generated sequences. Each motif occurrence is plotted at its genomic position, highlighting the combinatorial nature of motif clustering. The recurrence of canonical enhancer-associated motifs such as ETS1, SP1, and NFYA suggests that the generative model produces realistic cis-regulatory grammar beyond single-motif enrichment.

\newpage

\end{document}